\documentclass[letterpaper, 10 pt, conference]{ieeeconf} 
\makeatletter
\let\NAT@parse\undefined
\makeatother
\IEEEoverridecommandlockouts % for \thanks
\usepackage[colorlinks=true,linkcolor=blue,filecolor=magenta]{hyperref}
\usepackage{url}

\usepackage[ruled]{algorithm2e}

\let\labelindent\relax
\usepackage{enumitem}

\usepackage{mathtools}
\usepackage{amsmath}
\usepackage{amssymb}
\usepackage{bm}
\usepackage{xcolor}
\usepackage{multirow}
\usepackage{amsthm}

\newtheoremstyle{dense}%
  {3pt}% Space above
  {3pt}% Space below
  {\itshape}% Body font
  {}% Indent amount
  {\bfseries}% Theorem head font
  {:}% Punctuation after theorem head
  {.5em}% Space after theorem head
  {}% Theorem head spec (can be left empty, meaning `normal')
\theoremstyle{dense}
\newtheorem{remark}{Remark}
\AtBeginEnvironment{remark}{\small}

\DeclareMathOperator*{\argmin}{arg\,min}

\newcommand\best[1]{\colorbox{gray!33}{#1}}
\newcommand\better[1]{\colorbox{gray!12}{#1}}

\newcommand\seqone{\emph{Seq 1}}
\newcommand\seqtwo{\emph{Seq 2}}
\newcommand\seqthree{\emph{Seq 3}}
\newcommand\seqfour{\emph{Seq 4}}
\newcommand\seqfive{\emph{Seq 5}}
\newcommand\seqsix{\emph{Seq 6}}
\newcommand\seqseven{\emph{Seq 7}}
\newcommand\seqeight{\emph{Seq 8}}

\title{\LARGE \bf
BEVRender: Vision-based Cross-view Vehicle Registration in Off-road GNSS-denied Environment
}

\author{Lihong Jin, Wei Dong, Wenshan Wang, and Michael Kaess
\thanks{The authors are with the Robotics Institute, Carnegie Mellon University, PA, USA \texttt{\{lihongj, weidong, wenshanw, kaess\}@andrew.cmu.edu}}
\thanks{This material is based upon work supported by the U.S. Army Research Office and the U.S. Army Futures Command under Contract No. W911NF-20-D-0002.
    The content of the information does not reflect the position or the policy of the government, and no official endorsement should be inferred.
    }
}
\begin{document}

\maketitle
\thispagestyle{empty}
\pagestyle{empty}

%%% begin abstract
\begin{abstract}
%We introduce BEVRender, a novel learning-based approach for the localization of ground vehicles in Global Navigation Satellite System (GNSS)-denied off-road scenarios, which are typically challenging for conventional vision-based state estimation due to the lack of distinct visual landmarks and the instability of vehicle poses. 
We introduce BEVRender, a novel learning-based approach for the localization of ground vehicles in Global Navigation Satellite System (GNSS)-denied off-road scenarios. 
These environments are typically challenging for conventional vision-based state estimation due to the lack of distinct visual landmarks and the instability of vehicle poses. 
To address this,
BEVRender generates high-quality local bird's-eye-view (BEV) images of the local terrain. Subsequently, these images are aligned with a georeferenced aerial map through template matching to achieve accurate cross-view registration. 
Our approach overcomes the inherent limitations of visual inertial odometry systems and the substantial storage requirements of image-retrieval localization strategies, which are susceptible to drift and scalability issues, respectively.
Extensive experimentation validates BEVRender's advancement over 
%contemporary
existing GNSS-denied visual localization methods, demonstrating notable enhancements in both localization accuracy and update frequency. 
%
% The code for BEVRender will be made available soon.
% -> We will release the code upon acceptance.
\end{abstract}
%%% end abstract

\begin{figure*}[ht]
    \includegraphics[width=\textwidth]{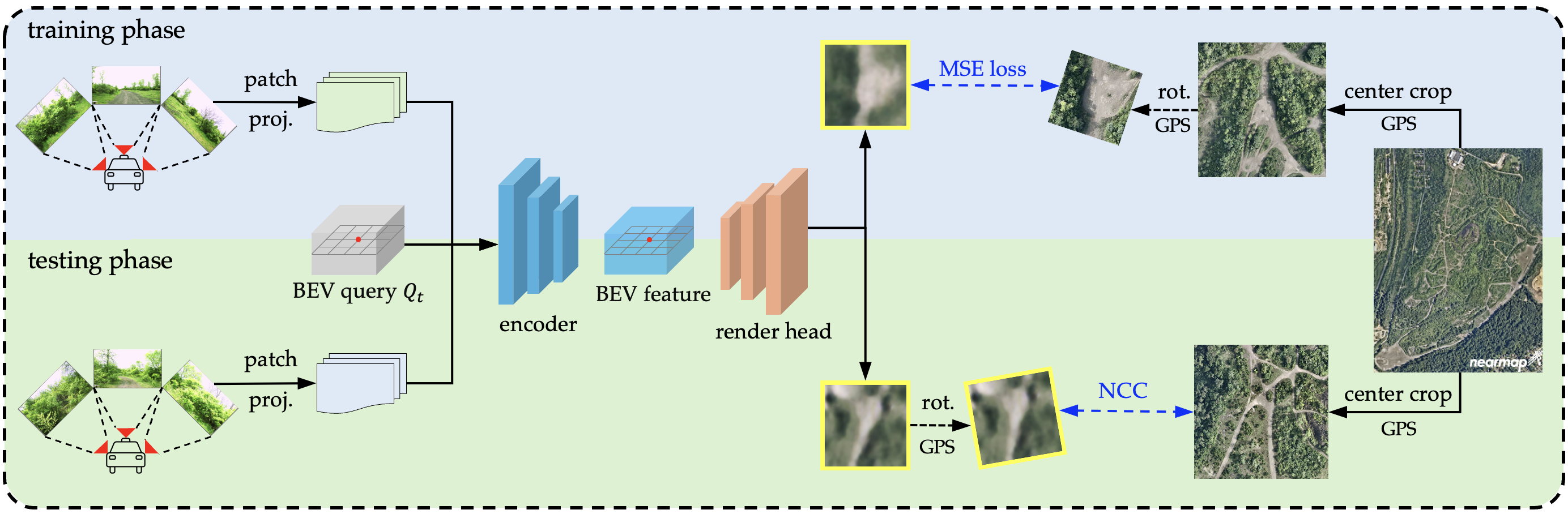}
    \caption{A diagram of our system. The light blue background indicates the training phase and the light green background indicates the testing phase. During the training phase, camera images are patch projected and sent to the feature encoder (in blue) and rendering head (in orange) to generate BEV images (highlighted in yellow boxes). The aerial map image is rotated and cropped according to the GPS information provided, ensuring that the final label image accurately represents the BEV space surrounding the vehicle. During the testing phase, the rendered BEV image is rotated according to the azimuth angle provided by the GPS, and matched against a local search region surrounding the vehicle position.}
    \label{fig:system}
\end{figure*}

\section{Introduction}

%Localizing ground vehicles in off-road conditions poses substantial challenges, primarily due to the lack of distinct visual landmarks, such as roads and buildings, which are crucial for vision-based state estimation. 
% It is challenging to localize ground vehicles in off-road non-urban environments such as forest, desert, and barren landscapes. A lack of distinct visual landmarks, such as roads and buildings, often raises difficulties for classical vision-based state estimation. 
Global localization is a crucial component that supports smooth navigation of autonomous vehicles. It is typical to equip on-board localization systems with the Global Navigation Satellite System (GNSS) modules for consistent and reliable global poses. 
However, in reality, GNSS signals can be blocked due to natural or artificial barriers, causing temporal system failures, where vision-based localization (VBL) serves as an alternative in GNSS-denied localization. 
A variety of methods have been proposed for VBL in urban scenarios~\cite{vbl_survey}, yet off-road VBL for unmanned ground vehicle (UGV) is still challenging due to non-urban environments lacking stable and distinct visual features, such as roads and buildings. 
The varied and unpredictable terrain further complicates the task by inducing unstable vehicle poses, making it difficult to maintain consistent feature matching across frames. 
% These challenges are amplified in environments where natural obstructions impede Global Navigation Satellite System (GNSS) signals, leading to potential system failures.
% \wenshan{should we highlight the large view-point difference between map and input images? } \wenshan{providing a visualization of the task might be helpful}

In response to these challenges, our paper presents a novel learning-based method that synthesizes a local bird's eye view (BEV) image of the surrounding area by aggregating visual features from camera images. This approach integrates a modified BEVFormer~\cite{li2022bevformer} framework with a novel rendering head, employing template matching for precise cross-view registration between ground vehicles and aerial maps in GNSS-denied off-road environments.

We concentrate on 2D relocalization of unmanned ground vehicles (UGV) for non-urban settings 
%such as forests, deserts, and barren landscapes, 
bounded within defined areas. Equipped with trinocular RGB cameras and an Inertial Measurement Unit (IMU), the vehicle employs multi-view visual inertial odometry (VIO) for state estimation. Our aim is to achieve accurate 2D positioning relative to a geo-referenced aerial map, facilitating pose correction in the absence of GNSS signals, whether temporarily or persistently. A more detailed problem definition is in Sec.~\ref{problem_definition}.

Previous study~\cite{Litman} has explored the creation of orthographic view images by accumulating geometric features over consecutive frames, coupled with Normalized Cross-correlation (NCC) for relocalization in a GPS-denied situation. However, this approach is limited by the inherent drift of VIO systems, which can distort the accumulated geometric data, leading to inaccuracies in ground-to-air matching. Our paper introduces a learning-based strategy for generating BEV images, using a Vision Transformer (ViT)~\cite{dosovitskiy2020vit}-based network for feature encoding. This method shows improved performance in generating local BEV images and supporting vehicle localization with geo-referenced aerial maps.
% \wenshan{Is there an insight about why using ViT to generate BEV images is a good idea? You just said VIO systems have large drift, is it the case the learning-based BEV map is more accurate? } \wenshan{Maybe we can also emphasize the difficulties of the off-road environment, which makes applying BEVFormer non-trivial. }

%Contrary to other research efforts~\cite{Burkhardt2005, sarlin2024snap, geodtr} that treat vision-based localization as an image retrieval problem, requiring substantial storage for on-board localization systems, our approach generates local BEV images for direct template matching. This significantly reduces the need for extensive data storage, relying instead on a geo-referenced map for real-time 2D localization. In summary, our contributions are fourfold:

Other research efforts~\cite{Burkhardt2005, sarlin2024snap, geodtr} treat vision-based localization as an image retrieval problem, requiring substantial storage for on-board localization systems. On the contrary, our approach generates local BEV images for direct template matching. This significantly reduces the need for extensive data storage, relying instead on a geo-referenced map for real-time 2D localization. In summary, our contributions are threefold:
% \wei{Experimental result is usually not a contribution... Maybe rethink and rephrase a bit and retain 3 items in the list.}

\begin{enumerate}[leftmargin=*]
    \item We propose a novel \emph{learning}-based framework for ground vehicle localization that combines BEV image generation with \emph{classical} template matching, eliminating the extensive dataset storage requirements of image-retrieval-based localization.
    % \item We introduce an efficient image rendering head as a feature decoder following the modified BEVFormer~\cite{li2022bevformer} encoder, capable of producing detailed top-down views of the local terrain.
    \item We integrate the deformable attention module in~\cite{Xia_2022_CVPR} with the BEVFormer network, enhancing feature encoding by using offset networks~\cite{Xia_2022_CVPR}, followed by an efficient image rendering head as a feature decoder capable of producing detailed top-down views of the local terrain.
    % \wei{Merge point 2 and 3.}
    \item Through comprehensive experiments with real-world datasets, we demonstrate that our method exhibits superior localization accuracy and frequency compared to existing GNSS-denied visual localization techniques, and generalizes to unseen trajectories.
\end{enumerate}

\section{Related Work}
\subsection{GNSS-denied Vehicle Localization}

Vehicle localization in GNSS-denied environments can be broadly categorized into relative and absolute localization strategies. Relative localization aims to mitigate odometry drifts by fusing data from multiple onboard sensors with motion models, or by leveraging loop closures to correct drift relative to global frames~\cite{hess2016real}. 
Absolute localization, in contrast, involves constructing local maps from the vehicle's perspective and aligning them with a global georeferenced map to determine precise vehicle positions. 
Reference data for this process can vary, including High-definition (HD) maps~\cite{poggenhans2018lanelet2}, aerial satellite imagery, Digital Elevation Models (DEM)~\cite{wan2022terrain,dem_localization}, and OpenStreetMap (OSM) data~\cite{sarlin2023orienternet}. While HD maps offer high accuracy, they are costly and data intensive. DEMs, primarily used for UAVs~\cite{wan2022terrain}, cater to non-planar terrains and scale ambiguity, whereas OSM provides dense semantic and geometric details suitable for urban navigation. Aerial satellite maps present strong visual cues with detailed information for off-road localization.

Significant advancements have been made in aligning ground-level images with aerial imagery for localization. Viswanathan et al.~\cite{huber2014iros} demonstrate effective ground-to-air image matching using satellite images by warping UGV panoramic images to a bird's eye view, comparing feature descriptors, and employing a particle filter for accurate localization. Based on this, recent work~\cite{Litman} focuses on generating an orthographic occupancy map by accumulation of local features and estimation of pose through NCC, and optimizing the prediction of global pose through a registration graph~\cite{factor_graphs_for_robot_perception}. In contrast, our approach adopts a Vision Transformer (ViT)-based~\cite{dosovitskiy2020vit} learning network to generate BEV images for ground-to-air matching, emphasizing frame-by-frame registration accuracy and reducing reliance on global trajectory optimization.

\subsection{Learning Vision-based Localization}

The evolution of vision-based localization has seen it conceptualized as an image retrieval task~\cite{Burkhardt2005}, employing contrastive learning to enhance the matching of onboard camera and satellite images~\cite{sarlin2024snap,geodtr}. Efforts to improve image alignment include warping satellite imagery by polar transformation to match ground perspectives~\cite{geodtr}, and constructing semantic neural maps from camera images~\cite{sarlin2024snap}. Further innovations leverage CNNs for feature extraction and BEV representation, enabling precise localization through 3D structure inference and matching~\cite{sarlin2023orienternet, camiletto2023ubev, fervers2023cbev, zhang2022bevlocator}.

The advent of foundation models offers promising directions for Visual Place Recognition (VPR), demonstrating the adaptability of pre-trained models (e.g., DINO~\cite{Caron_2021_ICCV}, DINOv2~\cite{oquab2024dinov2}) to diverse environments without fine-tuning~\cite{keetha2023anyloc}. Subsequent work~\cite{yao2023foundloc} integrates dense visual feature extraction with advanced filtering and global-local pose estimation via Extended Kalman Filters (EKF) for refined localization accuracy. Our methodology aligns with these advancements, utilizing a streamlined ViT architecture for efficient and accurate BEV image rendering and localization, minimizing parameter overhead while maximizing performance.

In the realm of self-driving applications, BEV representations~\cite{Can_2021_ICCV,bev_perception_review} have been enriched by encoding temporal and spatial features, as demonstrated by BEVFormer~\cite{li2022bevformer}, which leverages attention mechanisms~\cite{vaswani2017attention, Xia_2022_CVPR} for 3D object detection. Our work extends this concept by incorporating BEVFormer's feature propagation approach, ensuring our BEV representations integrate temporal information from successive frames. This strategy is complemented by recent explorations in temporal information encoding for BEV representation, highlighting the continuous evolution and application of these techniques in autonomous navigation~\cite{hu2023fusionformer,richard2021,Qin_2023_ICCV,cai2023bevfusion4d,stretchBEV}.
%Proposed in~\cite{Xia_2022_CVPR}, BEV

\begin{figure*}[ht]
\includegraphics[width=\textwidth]{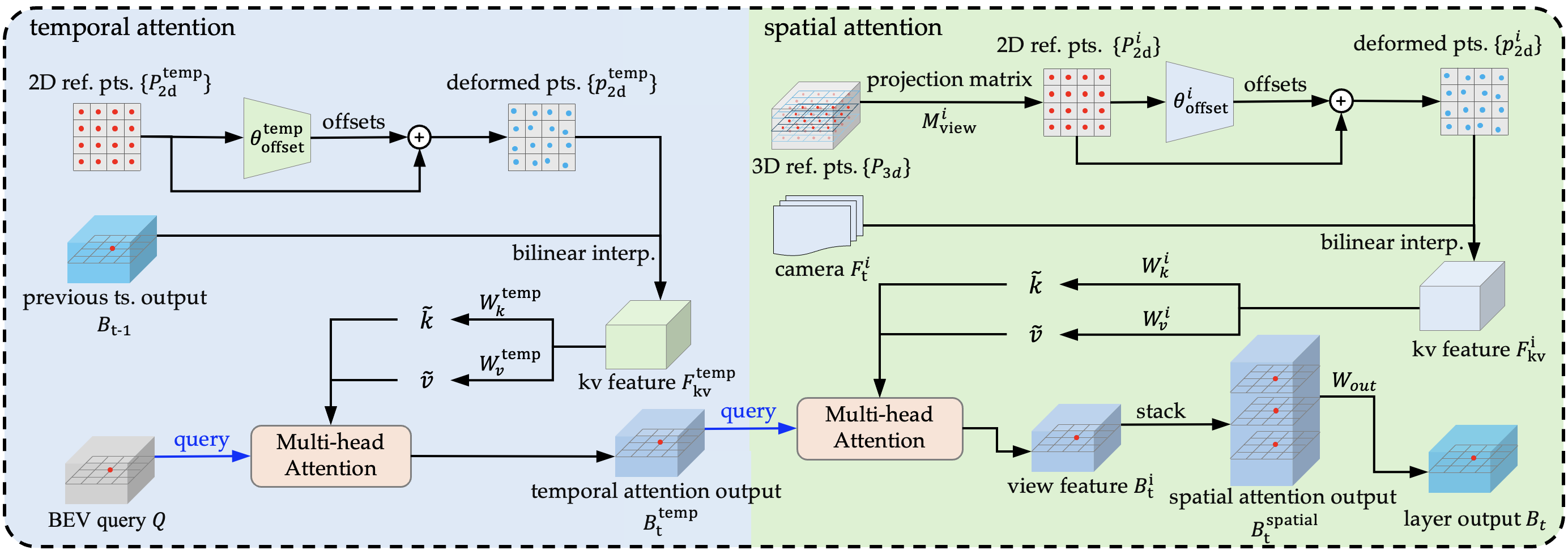}
\caption{Encoder layer architecture. An encoder layer is composed of temporal and spatial attention. In temporal attention, a set of 2D reference points with a spatial dimension of $l\times w$ is sampled and deformed. Next, bilinear sampling is performed to extract tokens for multi-head attention (MHA)~\cite{vaswani2017attention} given deformed reference points from previous timestamp BEV feature $B_{t-1}$. The MHA output from temporal attention serves as a query for the subsequent spatial attention module. In spatial attention, we sample one point per 3D grid cell in the BEV space as reference points and project them to the three camera image frames with extrinsic and intrinsic parameters to obtain 2D reference points for each image view. Similarly to temporal attention, the 2D reference points are deformed and used for bilinear sampling, but from camera feature. A more detailed description can be found in Sec.~\ref{encoder_layer_description}.}
\label{fig:approach-arch}
\end{figure*}

\section{Methodology}\label{problem_definition}
Our system contains three main components: a feature encoder to map the visual features from the camera to the top-down view, a rendering head to decode the features and generate top-down BEV images, and an image registration component for cross-view localization. An overview of our system is shown in Fig.~\ref{fig:system}.

We consider a scenario where a vehicle, equipped with trinocular cameras and an IMU, is traversing flat natural terrain. A pre-stored aerial map of the area aids in localization. The vehicle's pose is predicted by the VIO system in a local coordinate frame as follows:
\begin{align}
        \mathbf{X}_t = \Big[x_t, y_t, \theta_t\Big] \in \mathbf{SE}(2).
\end{align}

We assume that the prediction for the azimuth angle $\theta_t$ from VIO is accurate, but the position estimates ($x_t$ and $y_t$) may drift over time. Our system aggregates information from consecutive frames to construct a top-down representation of the environment for map registration.

We define a 3D BEV space centered on the vehicle with a length of $L$, a width of $W$ and a height of $H$. The space is divided into $l\times w\times h$ grid cells, so that each cell represents a cubic size of $\frac{L}{l}\times\frac{W}{w}\times\frac{H}{h}$ in the real world. The BEV query is a 3D trainable embedding with a dimension of $l\times w\times h$ representing the BEV space and serving as the query for deformable attention modules in the encoder. All intermediate BEV features in the network also follow the same spatial dimension. The specific range and dimension chosen for our experiment are described in Sec.~\ref{experiment_setting}. 

Our system seeks to find the optimal pose prediction that minimizes the difference between camera feature representation and local aerial image:
\begin{align}
%    \min_{\mathcal{X}^*\in\{\mathcal{X}\}}\psi\Big(I^{'}_{\text{bev}}(\mathcal{X}^*), I_{\text{map}}(\mathcal{X}^*)\Big),
\mathbf{X}^* = \argmin_{\mathbf{X}}\Phi\Big(I^{'}_{\text{bev}}(\mathbf{X}), I_{\text{map}}(\mathbf{X})\Big),
\end{align}
where $\Phi$ is a function to find $\mathbf{X}^*$ to achieve minimum distance between two representations in feature space, and is provided by template matching module in our system. $I_{\text{map}}$ is a subset of the aerial map with respect to a vehicle pose. 

The image rendering head $\Psi_\text{render}$ defines a mapping from the encoded image feature $F_{\text{feat}}$ to top-down BEV image $I^{'}_{\text{bev}}$:
% and $I^{'}_{\text{feat}}$ is the rendered BEV image given learned feature:
\begin{align}
    I^{'}_{\text{bev}}(\mathbf{X}) = \Psi_{\text{render}}\Big(F_{\text{feat}}(\mathbf{X})\Big).
\end{align}
% where $\Psi_\text{render}$ is a mapping from the encoded feature $F_{\text{feat}}$ to top-down BEV image, given by the rendering head.

\subsection{Feature Encoding with BEVFormer}\label{encoder_layer_description}
Adopting BEVFormer's framework~\cite{li2022bevformer}, we propagate consecutive frame features to capture temporal information. Within a temporal window of $T$ seconds, $n$ frames ($3\times n$ images in a trinocular setup) are sampled. A detailed setting can be found in Sec.~\ref{experiment_setting}.
Starting with the earliest frame, camera images $I^{\mathrm{cam}}_{t}$ are processed through patch projection, which is a convolutional layer in our implementation, to obtain camera feature $F^{\mathrm{cam}}_{t}$
and sent to the encoder together with the BEV query $Q$ and previous BEV feature $B_{t-1}$ to obtain the encoded BEV feature for current timestamp $B_t$. The encoding process consists of two stages: a temporal attention stage that takes in query $Q$ and previous timestamp BEV feature $B_{t-1}$ for deformable attention:
\begin{align}
    B^{\text{temp}}_{t} = \text{DeformableAttn}\Big(B_{t-1}, Q\Big),
\end{align}
followed by a spatial attention stage that takes in temporal output and camera feature $F_{t}$ for deformable attention:
\begin{align}
    B_{t} = B^{\text{spatial}}_{t} = \text{DeformableAttn}\Big(F^{\text{cam}}_{t},B^{\text{temp}}_{t}\Big).
\end{align}

$B_{t}$ is then projected to the subsequent frame vehicle pose
as $B^{'}_{t}$ according to the movement of the vehicle provided by GPS. The projection is performed by affine transformations in $\mathbf{SE}(2)$ followed by a bilinear interpolation:
\begin{align}
    \Delta \mathbf{X} = \mathbf{X}_t - \mathbf{X}_{t-1}= \Big[\Delta x, \Delta y, \Delta \theta\Big] ,
\end{align}
\begin{align}
    \left[\begin{matrix}
        x_t\\y_t\\1
    \end{matrix}\right]=\left[\begin{matrix}
        \cos{\Delta\theta} & -\sin{\Delta\theta}& \Delta x\\
        \sin{\Delta\theta} & \cos{\Delta\theta}& \Delta y\\
        0&0&1
    \end{matrix}\right]\left[\begin{matrix}
        x_{t-1}\\y_{t-1}\\1
    \end{matrix}\right],
\end{align}
\begin{align}
    B^{'}_{t}(x_t, y_t) = \mathrm{BilinearInterp} (B_t(x_t,y_t)).
\end{align}

\begin{figure}[ht]
\includegraphics[width=\columnwidth]{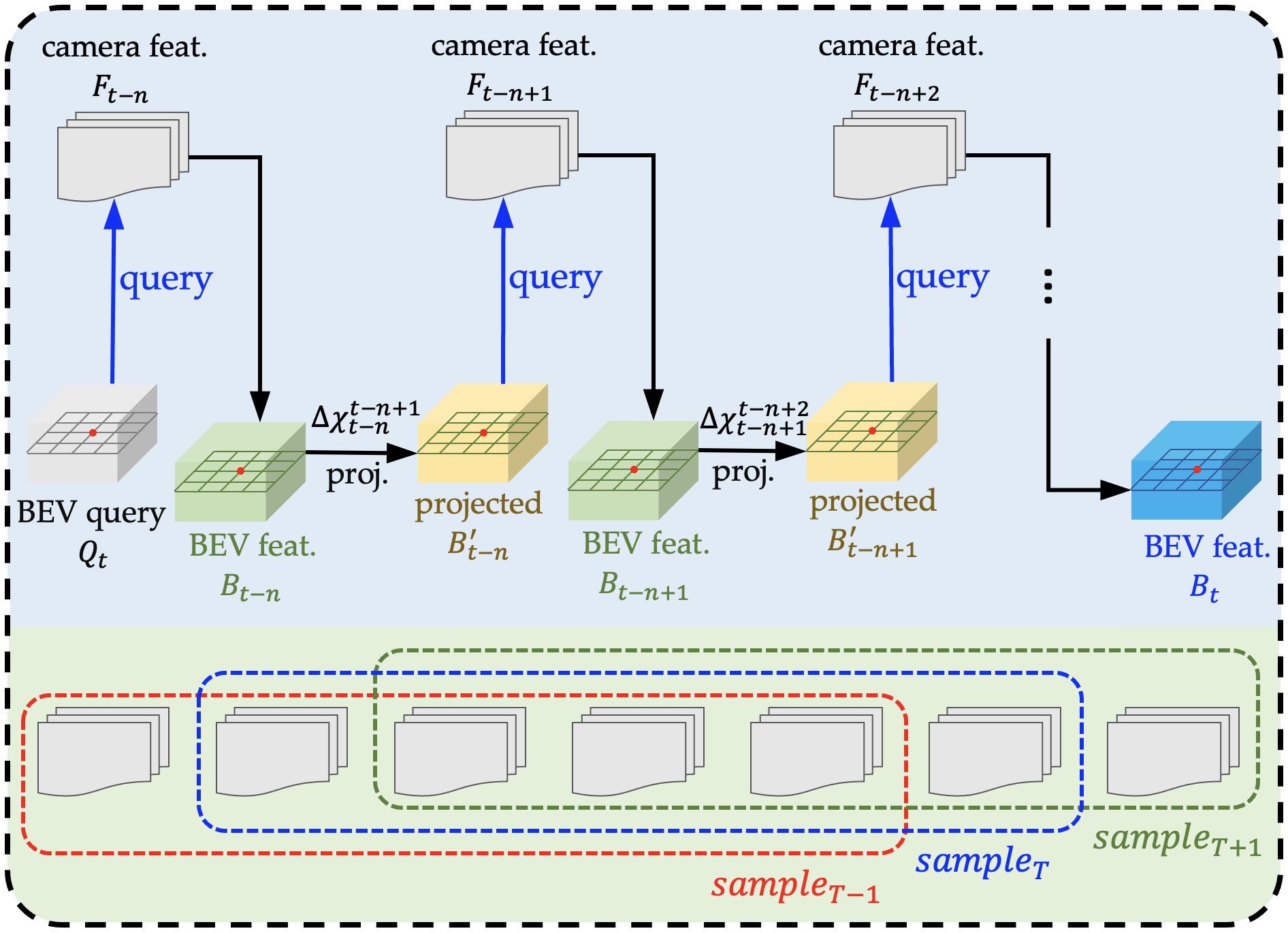}
\caption{Temporal feature propagation and dataset organization. For each timestamp, we sample $n$ frames from past $T$ seconds, composing a training sample of $n\!+\!1$ camera frames together with current timestamp frame. Staring with the earliest timestamp in the window, BEV query $Q$ is used to query camera feature $F$ to obtain BEV feature $B$, which is subsequently projected to next timestamp vehicle position given GPS outputs, to obtain new feature $B'$. Propagation continues until the latest frame is processed. A detailed description on projection can be found in Sec.~\ref{encoder_layer_description}.
}
\label{fig:approach-detail}
\end{figure}

Subsequently, $B^{'}_{t}$ serves as a query to the encoder to query the next timestamp camera feature $F_{t+1}$ to obtain $B_{t+1}$. The propagation continues in the temporal window until we obtain the latest timestamp feature. A diagram of temporal propagation is shown in Fig.~\ref{fig:approach-detail}. It should be noted that $B_{t-1}$ is the same as query $Q$ for the first frame in the temporal window:
\begin{align}
    B_{t-1}=Q\quad \text{if}\quad t=0.
\end{align}

Unlike BEVFormer, our encoder simplifies to a single layer, totaling 1.44 million parameters, while supporting effective feature learning for downstream localization tasks. 

The architecture of the encoding layer is shown in Fig.~\ref{fig:approach-arch}, and the ablation study of the number of layers can be found in Table~\ref{tab:ablation}.

\subsection{Deformable Attention Vision Transformer}
In contrast to BEVFormer that employs Deformable DETR~\cite{zhu2020deformable},
%for its deformable attention module, 
our approach utilizes the deformable attention 
%proposed in
~\cite{Xia_2022_CVPR}, which uses offset networks to calculate adjustments to each reference point. The offsets are processed by an additional convolution layer $\theta_{\text{offset}}$, as shown in Fig.~\ref{fig:approach-arch}, and its output modifies the original reference point to generate deformed reference points. For spatial attention, offsets $\theta^i_{\text{offset}}$ are added to the reference points unique to each camera view $i$, acting as adjustments to the pixel locations of reference points. Consequently, we employ three distinct convolution layers dedicated to learning offsets as an adaptation to the trinocular system setting. The final output of the spatial attention layer is a stacking of features from three camera views, undergoing another convolutional layer to maintain the same spatial dimension as the BEV query and BEV features. 

%\wei{Might be better to remove this equation due to a lack of context; Otherwise you need to describe all the variables.}.
%\lihong{are we removing this?}

The output of deformable attention heads is formulated as
\begin{align}
    z^{(m)} = \sigma\Big( \frac{q^{(m)}{k^{(m)}}^\top}{\sqrt{d}} +\phi(B;R) \Big) v^{(m)},
\end{align}
where $q, k, v$ constructs the standard transformer attention~\cite{vaswani2017attention} with softmax activation $\sigma$ and scale normalization $\sqrt{d}$, enhanced by relative positional bias~\cite{liu2021Swin} in $\phi(B;R)$. A more detailed description of deformable attention formulation can be found in~\cite{Xia_2022_CVPR}.

\subsection{BEV Image Rendering Head}
The BEV image rendering head is designed to translate encoded features into interpretable top-down views of the vehicle's surroundings. It is a straightforward convolutional neural network (CNN) architecture that takes as input the encoded BEV features with dimensions of $d\times l\times w$, where $d$ is the model embedding dimension. Through a series of convolutional and upsampling layers, the BEV features are processed to generate a colored image of certain size, which serves as a top-down visual representation of the BEV space around the vehicle. The rendering head ensures that the resulting BEV image retains critical spatial information required for ground-to-aerial vehicle localization in GNSS-denied environments. The detailed structure of the rendering head is illustrated in Table~\ref{tab:render_head}.

\begin{table}[ht]
\centering
\caption{BEV rendering head architecture}
\resizebox{.48\textwidth}{!}{
\begin{tabular}{c| c}
\hline
block & layer\\
\hline\hline
Decoder block 0 & Conv2d + BN + ReLU\\
Decoder block 1 & (Conv2d + BN)$\times$4 + ReLU\\
Decoder block 2 & (Conv2d + BN)$\times$4 + ReLU  \\
Decoder block 3 & (Conv2d + BN)$\times$4 + ReLU  \\
Upsample block 0 &  Upsample + (Conv2d + BN)$\times$2 + ReLU\\
Upsample block 1 &  Upsample + (Conv2d + BN)$\times$2 + ReLU\\
Upsample block 2 &  Upsample + (Conv2d + BN)$\times$2 + ReLU\\
Upsample block 3 &  Upsample + (Conv2d + BN)$\times$2 + Sigmoid\\
\hline
\end{tabular}
}
\label{tab:render_head}
\end{table}

\begin{figure*}[ht]
    \centering
    \resizebox{\textwidth}{!}{
    \begin{tabular}{@{}c@{\hspace{1mm}}c@{\hspace{1mm}}c@{\hspace{1mm}}c@{}}
        \includegraphics[width=.24\textwidth]{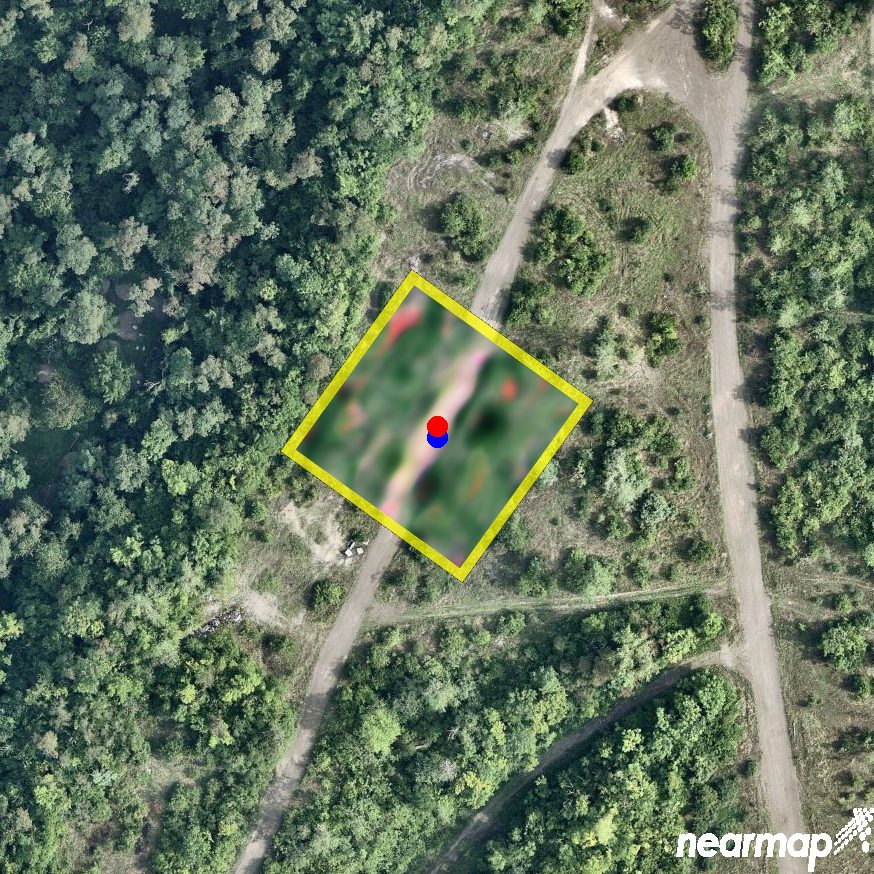} &
    \includegraphics[width=.24\textwidth]{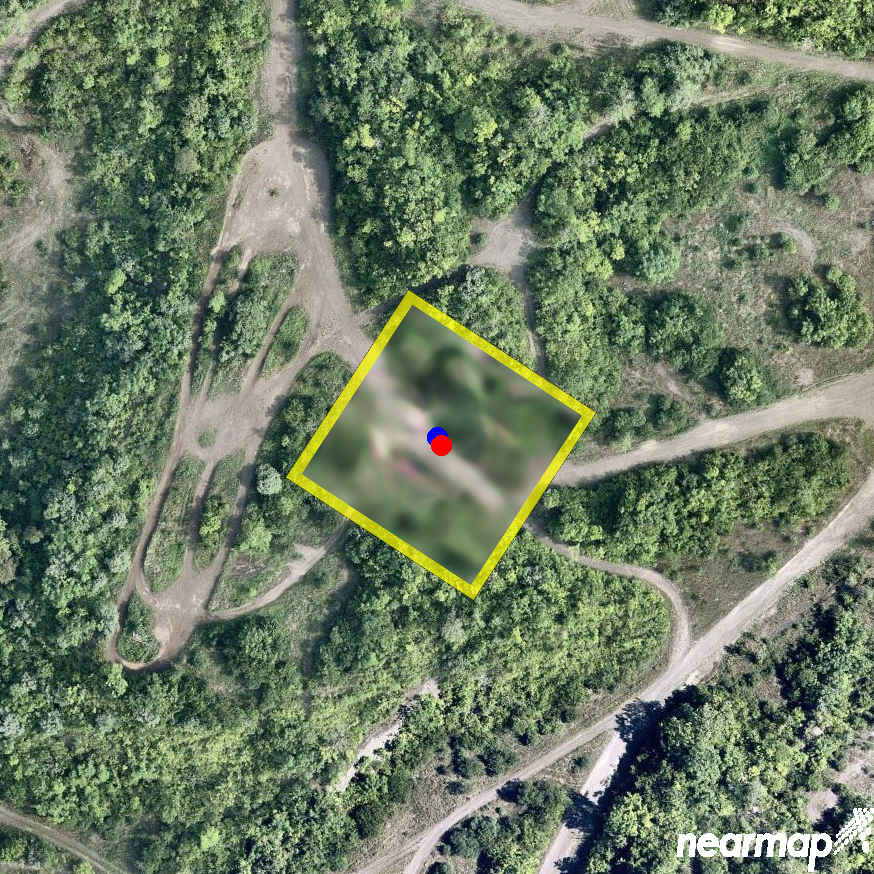} &
    \includegraphics[width=.24\textwidth]{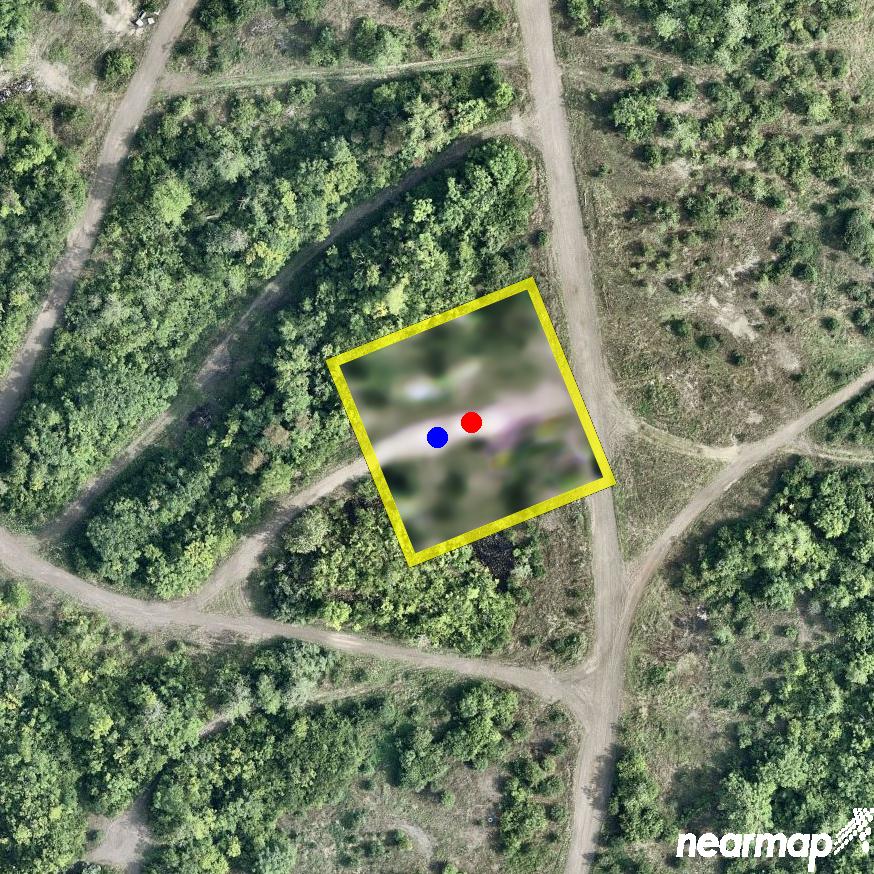} &
    \includegraphics[width=.24\textwidth]{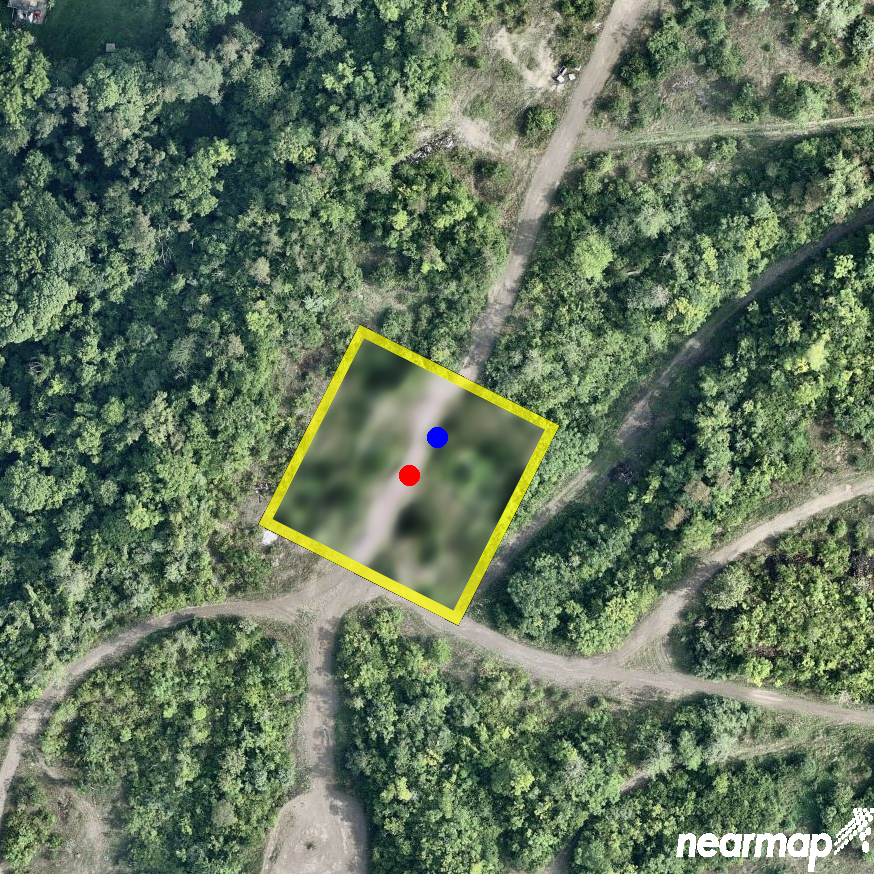}\\
    \includegraphics[width=.24\textwidth]{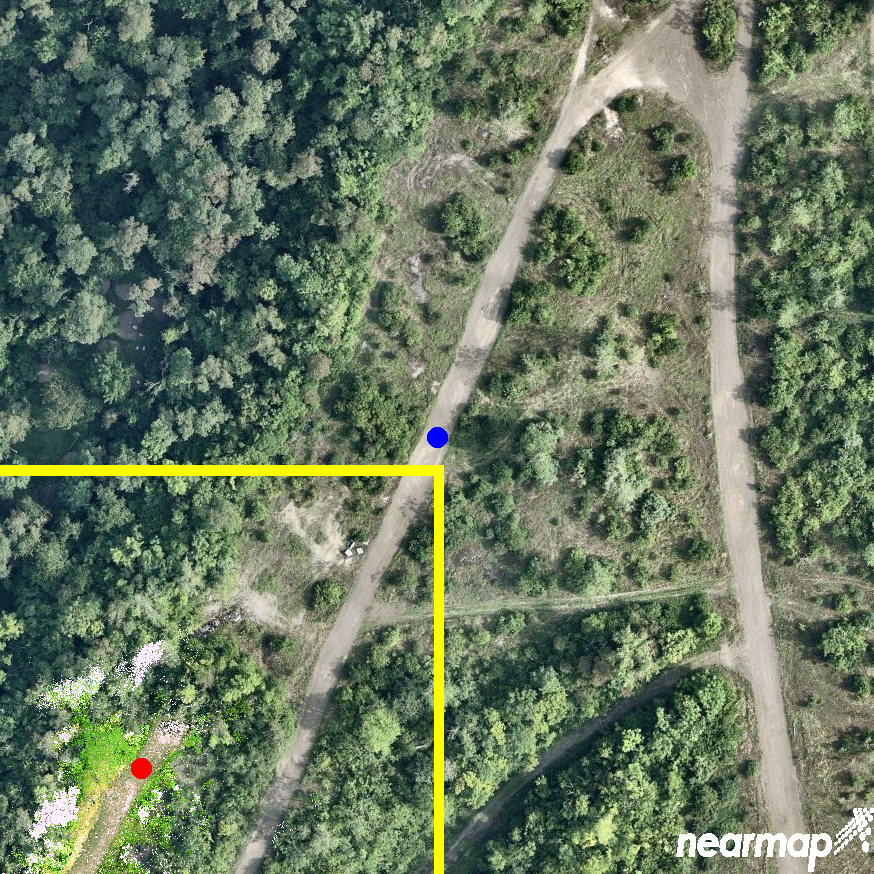} & 
    \includegraphics[width=.24\textwidth]{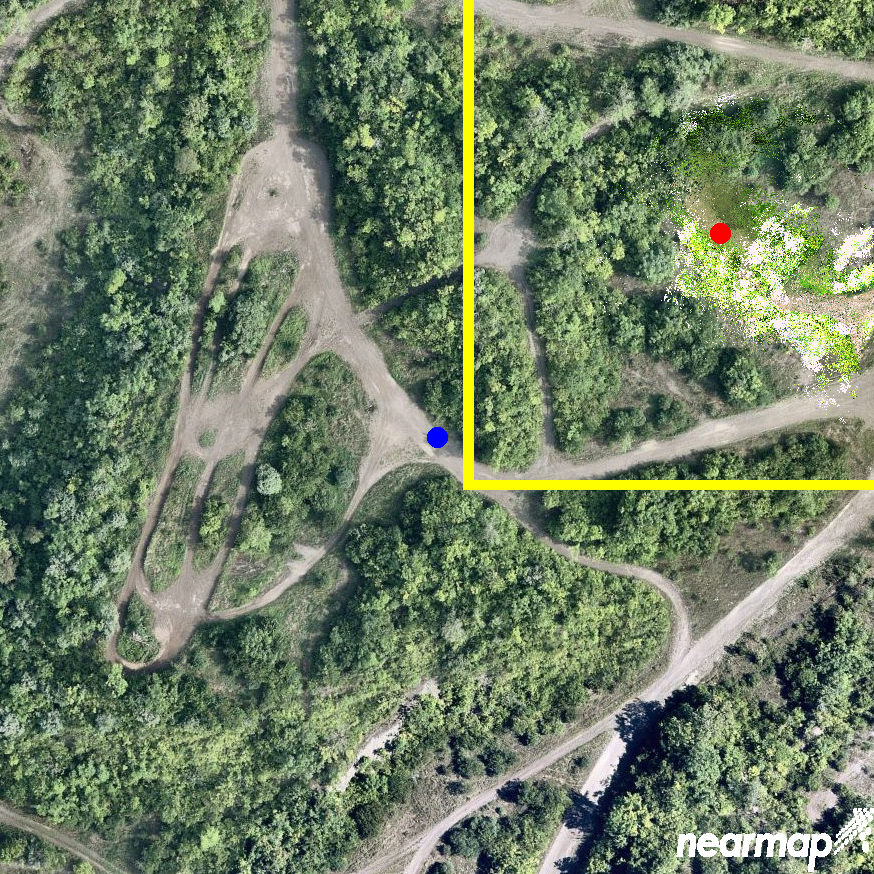} &
    \includegraphics[width=.24\textwidth]{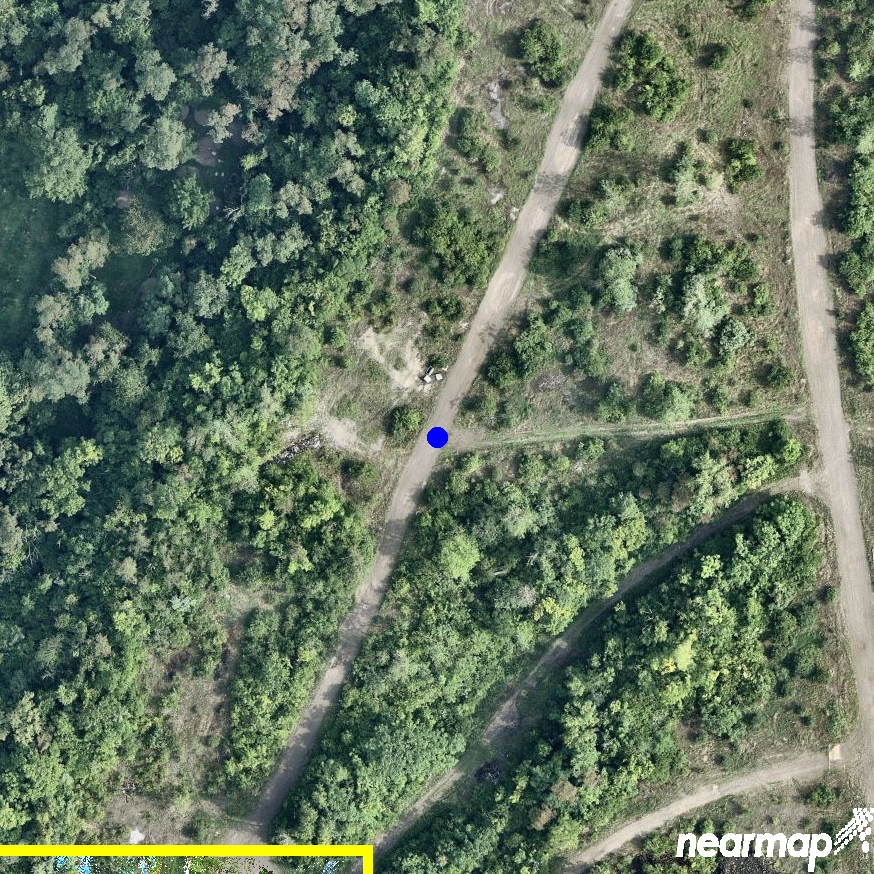} &
    \includegraphics[width=.24\textwidth]{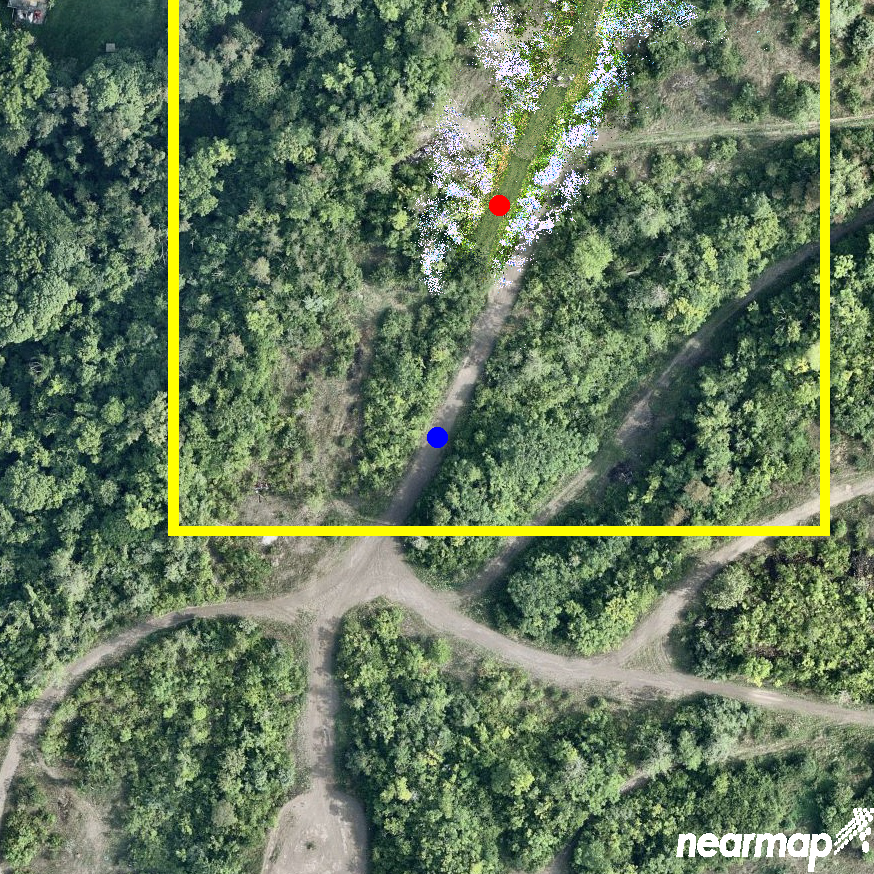}\\
    \footnotesize{\seqone} & \footnotesize{\seqtwo} & \footnotesize{\seqthree} & \footnotesize{\seqfour}
    \end{tabular}
    }
    \caption{Qualitative comparison of our method and Litman~\cite{Litman}. 
    \emph{Top row}: the rendering and registration result of our method, where the BEV images are highlighted in yellow boxes, the red dots indicate the NCC predictions from our system, and the blue dots indicate the GPS ground truth position. Our approach produces coherent rendering to the aerial image.
    \emph{Bottom row}: predictions from Litman~\cite{Litman}. Similarly, the red and blue dots indicate the predictions and ground truth, while the yellow boxes indicate the generated occupancy image overlaid on the groundtruth. Only semi-dense rendering are available for Litman~\cite{Litman} (see the saturated white and green points around red dots), resulting in compromised registration accuracy.
    }
    \label{fig:qualitative}
\end{figure*}

\begin{table}[ht]
\centering
\tiny
\caption{Statistics of GNSS-denied real-world dataset}
\resizebox{.48\textwidth}{!}{
\begin{tabular}{c|c|c|c}
\hline
& \# images & traj. length ($\mathrm{m}$) & coverage ($\mathrm{m}^2$)\\
\hline\hline
\seqone &  1634 & 1059.42 & 349.34$\times$159.70 \\
\seqtwo &  1563 & 1067.08 & 349.34$\times$159.67 \\
\seqthree &  1427 & 1415.72 & 353.07$\times$164.65 \\
\seqfour &  1210 & 1228.61 & 350.99$\times$161.92 \\
\seqfive & 1707 & 1179.64& 462.53$\times$359.25\\
\seqsix & 838& 495.64&340.13$\times$239.08\\
\seqseven &815 &439.67 & 410.86$\times$74.74\\
\seqeight & 1395& 1425.88&368.51$\times$166.06\\
\hline
aerial map&-&-&1278.20$\times$1646.46\\
\hline
\end{tabular}
}
\label{tab:dataset}
\end{table}

\section{Experiments}

% grid 916 seems to have better result 
% seq 1 - 19.33, 26.09, 63.44
% seq 2 - 22.40, 27.92, 60.90
% seq 3 - 20.60, 24.96, 58.93
% seq 4 - 21.18, 25.49, 57.50
\begin{table*}[ht]
    % \centering
    \caption{quantitative comparisons on our real-world dataset}    
    \centering
    \resizebox{\textwidth}{!}{
        \begin{tabular}{c *{4}{|c|c|c} }
        \hline
        \multirow{2}{*}{approach} & 
        \multicolumn{3}{c|}{\seqone} & 
        \multicolumn{3}{c|}{\seqtwo} & 
        \multicolumn{3}{c|}{\seqthree} & 
        \multicolumn{3}{c}{\seqfour} \\
        \cline{2-13} & mean $\downarrow$& std $\downarrow$ & match (\%) $\uparrow$ 
                     & mean $\downarrow$& std $\downarrow$ & match (\%) $\uparrow$ 
                     & mean $\downarrow$& std $\downarrow$ & match (\%) $\uparrow$                  
                     & mean $\downarrow$& std $\downarrow$ & match (\%) $\uparrow$ \\
        \hline\hline
        Litman~\cite{Litman} & \better{24.35}&  \best{13.50} & \better{21.62}(Rmk.\ref{remark:litman_match_rate}) & 
        \better{34.45} & \best{21.59} & \better{12.12}
        & \better{26.27} & \best{13.44} & \better{11.46}
        & \better{61.04} & 55.80 & \better{8.89} \\
        GeoDTR~\cite{geodtr} (top 1) & 82.72 & 25.52 &0.00 (Rmk.\ref{remark:gedtr_match_rate}) & 90.27 & 29.92 &1.28 & 84.40 & 27.33 &0.36 & 86.53 & \better{27.60} &0.00\\
        GeoDTR~\cite{geodtr} (top 5 avg.) & 67.35 & \better{24.22} &0.94 & 74.06 &  30.43 &1.60 &71.33 & 28.91&1.07 & 66.34 & 29.35 &1.67 \\
        \hline
         Ours & \best{19.33} & 26.09 & \best{63.44} 
              & \best{22.40} & \better{27.92} & \best{60.90}
              & \best{20.60} & \better{24.96} & \best{58.93}
              & \best{21.18} & \best{25.49} & \best{57.50}\\
         \hline
         \noalign{\smallskip}
        \end{tabular}
    }
     {\raggedright 1. The darker shading indicates the best results, and the lighter shading indicates the second-best results. \par
     \raggedright 2. The mean and std are calculated for the APE for predicted positions, see the registration accuracy of Sec.~\ref{registration_accuracy} for more details.\par
     \raggedright 3. The search region is set to 200$\times$200 square meters.\par}
    \label{tab:main}
\end{table*}

\subsection{Experiment Setting}\label{experiment_setting}

Since the satellite image\footnote{The satellite image used in this paper is provided by \href{https://www.nearmap.com/us/en}{Nearmap}.} has a resolution of 0.229 meters per pixel, we define the length and width of the BEV space as 25.648 meters centered on vehicle position, equivalent to a size of $112\times112$ pixels on the aerial map. We also define the height of the BEV space as 2 meters. The space is divided into $28\!\times\!28\!\times\!5$ 3D grid cells, so that each cell represents a voxel of $0.916\!\times\!0.916\!\times\!0.4\:m^3$ in the real world. We utilize a temporal window of 5 seconds and randomly sample 5 frames in the window to compose a training sample.

\begin{figure}[ht]
    \centering
    \resizebox{\columnwidth}{!}{
    \begin{tabular}{@{}c@{\hspace{1mm}}c@{\hspace{1mm}}c@{}}
    \includegraphics[width=.24\textwidth]{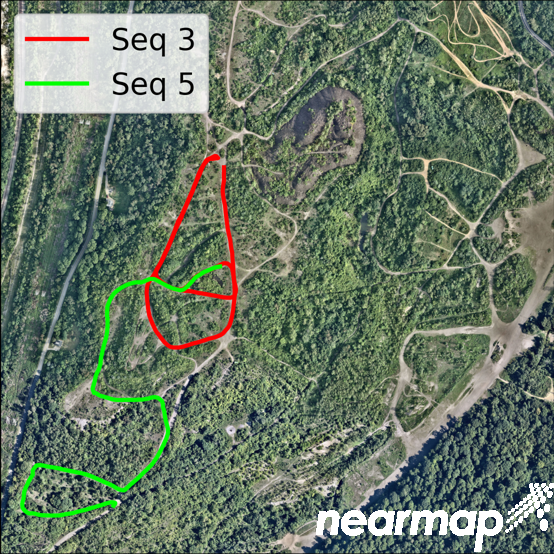} &
    \includegraphics[width=.24\textwidth]{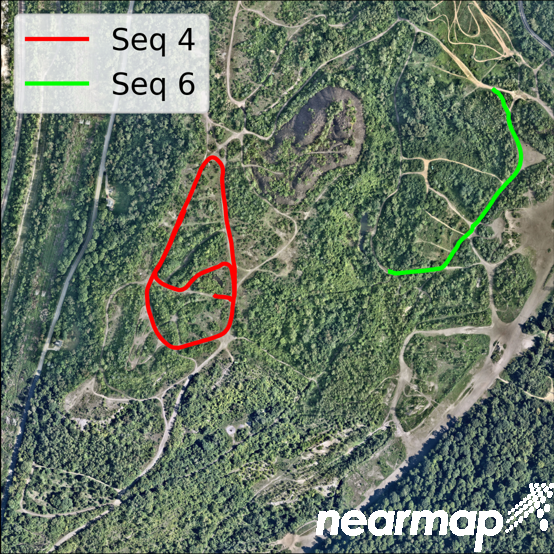} &
    \includegraphics[width=.24\textwidth]{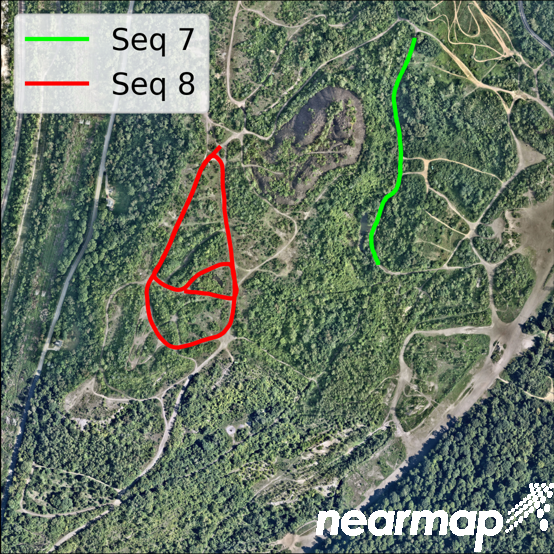}
    \end{tabular}
    }
    \caption{Trajectory plot for cross-sequence testing. Sequence 3 and 8 are used in training, sequence 4 to 7 are used in testing.}
    \label{fig:trajectory}
\end{figure}

We conduct two main experiments, one to compare against state-of-the-art VBL methods in GNSS-denied setting~\cite{Litman,geodtr}, where we use 4 sequences and split them into 80\% training, 20\% testing data; and another to show our model's ability to generalize across different scenes given limited training data, where we use 2 sequences for training and 4 sequences for testing. The trajectory plots for sequences used in the cross-sequence testing experiment are shown in Fig.~\ref{fig:trajectory}.

Training is distributed on 8 NVIDIA A100 GPUs for a total of 2500 epochs and with a learning rate of $4e^{-5}$. The configuration of the testing computer is described in Sec.~\ref{system_runtime} in system runtime.

During the testing phase, we crop and rotate the aerial map based on the GPS ground truth position as the center of the image with a size of 874$\times$874 pixels, which corresponds to a real-world coverage of approximately 200$\times$200 square meters. This search region is sufficient to accommodate VIO drift for more than 10 minutes without registration. For cross-sequence testing, we loosen the assumption of drifting range and use a search region of 100$\times$100 square meters. Our camera system captures 3 frames per second and predicts registration consistent with camera frame; therefore, sufficient for preventing from failing with the 100$\times$100 square meter search range.

%For template matching, we utilize the function \textit{matchTemplate} offered by OpenCV~\cite{opencv_library} and choose the method \textit{CCORR\_NORMED} in implementation. 
We use NCC for template matching. NCC identifies the best match within the search area, maximizing similarity between the generated BEV image and the aerial map, thus predicting the vehicle's position relative to the aerial map. We observe failure cases where rendered BEV images are of moderate visual quality, whereas NCC fails in prediction. An example of failure cases is shown in Fig.~\ref{fig:failure}.
% \ray{settings for NCC? Explain Fig 6.?}

\subsection{Dataset Organization}
We collect our real-world data set in the Pittsburgh area, with a VIO system on board. Detailed information on the sequences can be found in Table~\ref{tab:dataset}. For each training sample, we use the information of timestamp, trinocular RGB images, and GPS ground truth including x, y, and azimuth angle in the UTM coordinate system for training. The preprocessing for cropping the aerial map can be found in Fig.~\ref{fig:system}.

\subsection{Quantitative Comparison}

We compare our method with GPS denied registration via occupancy mapping proposed in~\cite{Litman}, and GeoDTR proposed in~\cite{geodtr}. The comparison result is shown in Table~\ref{tab:main}. 

Since GeoDTR is an image-retrieval-based method and relies on cultivating the corresponding information between camera inputs and polar transformation of aerial map images, it is required to preserve a database of candidate polar transformed images for real-world vehicle localization. We randomly sample 5000 particles within the search region at each timestamp and apply polar transformation according to the particle location on the map together with the azimuth angle of GPS ground truth. After obtaining the candidate polar images, for each timestamp, we pass in the camera images and polar images to the model, and calculate the distance between camera descriptors and polar descriptors, we choose candidate with closest descriptor distance as the top 1 prediction, and its corresponding real-world location as top 1 location, and we average the top 5 predicted locations as top 5 prediction.

\noindent \textbf{Registration accuracy}\label{registration_accuracy}
To evaluate the accuracy of vehicle registration, we calculate the mean and standard deviation (STD) of absolute position error (APE) between predicted position and the ground truth vehicle location given by on-board GPS. 

\noindent \textbf{Registration frequency}
In the real-world localization scenario, the update frequency is another important factor that determines the stability of registration system. We report the matching frequency by counting the total successful matches when the APE is within a threshold of 10 meters (the range we deem tolerable for our VIO system) and calculate the match rate as total successful matches divided by total camera frames for a sequence:
\begin{align}
    \mathbf{x}^i& = (x_i, y_i),\\
    \mathbf{d}^i&= \lVert \mathbf{x}^i_{\mathrm{gt}} - \mathbf{x}^i_\mathrm{{pred}} \rVert_2,\\
    \mathbf{p}_{\mathrm{match}} &= \frac{1}{N}\sum_{i=1}^N \mathbf{1}\boldsymbol{\cdot}(\mathbf{d}^i< \mathbf{d}_{\mathrm{threshold}}),
\end{align}
where $N$ is the number of images for a sequence per camera module.

\begin{remark}[Testing with Litman~\cite{Litman}]\label{remark:litman_match_rate}
It should be noted that the method proposed in~\cite{Litman} accumulates geometric features on a certain number of consecutive camera frames (50 by default), leading to a limited number of registration try-outs throughout a sequence. For comparison sake, we calculate the match rate as the total number of successful matches divided by the total number of occupancy maps synthesized in a sequence.
\end{remark}

\begin{remark}[Testing with GeoDTR]\label{remark:gedtr_match_rate}
It takes up to 21 hours to sample polar images for 5000 particles for 320 testing samples; hence, we cannot further increase the density of particles. 
To apply image-retrieval-based method for on-board localization, it is required to have a pre-stored dataset, specifically in our case, of polar images sampled from all candidate positions on local aerial map enumerating all possible rotations, which is prohibitively expensive storage for on-board system in real-world localization.
\end{remark}

\noindent \textbf{System runtime}\label{system_runtime} Testing is performed on a machine equipped with an AMD Ryzen 9 5900X 12-Core processor and a NVIDIA GeForce RTX 4090. The total time to localize 280 testing samples is 33.32 seconds, equivalent to 0.12 seconds to localize per camera frame. The camera frame rate for our system is 3 per second; therefore, our system is able to support online localization in real-world scenario.

\begin{table}[ht]
    \centering
    \tiny
    \caption{cross-sequence testing for model generalization}
    \resizebox{.35\textwidth}{!}{
    \begin{tabular}{c|c|c|c}
    \hline
    sequence & mean $\downarrow$ & std $\downarrow$ & match(\%) $\uparrow$\\
    \hline\hline
     \seqfour &11.24 & 6.64& 45.38\\
     \seqfive& 13.77&6.74 &31.16 \\
     \seqsix & 12.72& 6.38& 36.63\\
     \seqseven & 16.30&6.92 & 21.81\\
     \hline
    \end{tabular}
    }
    \label{tab:cross-sequence}
\end{table}

\subsection{Qualitative Comparison}
Visualizations of the rendering and registration result can be found in Fig.~\ref{fig:qualitative}. The image rendering head processes the encoded BEV feature with a spatial dimension of $64\times28\times28$ through a set of convolutional layers and 4 upsample layers, as shown in Table~\ref{tab:render_head}. The final BEV image is an RGB image with a size of $224\times224$ pixels, representing an area of $51.296\times51.296 \: m^2$. The occupancy map reconstructed from~\cite{Litman} aggregates geometric features from 50 consecutive frames, of which the coverage may vary for each prediction.

% \wei{Fig 3 and 5 are not referred to yet.}

\subsection{Model Generalization}
To test the generalizability of the proposed system, we perform cross-sequence tests. Specifically, training with sequences 3 and 8 while testing with sequences 4-7. We report search regions of $100\times 100 \: m^2$ in Table~\ref{tab:cross-sequence}.

\subsection{Ablation Study}
In this section we explore the influence of choosing different hyperparameters and BEV space resolutions on the final registration result. Since the aerial map resolution is 0.229 meters, we experiment with the BEV grid resolutions of 0.458 meters and 0.916 meters, corresponding to 2 pixels and 4 pixels on the map, respectively. We also experiment with an increased number of layers and report the results in Table~\ref{tab:ablation}. Considering the result from ablation study, we choose the resolution of the BEV grid as 0.916 meters, and the number of encoder layers as 1 for Table~\ref{tab:main} and Table~\ref{tab:cross-sequence}.

\begin{table}[ht]
    \centering
    \caption{Ablation study on Seq 4\\effects of architecture choice and hyper parameters}
    \resizebox{.48\textwidth}{!}{
    \begin{tabular}{c|c|c|c|c|c}
    \hline
    \# layers& grid reso. ($m$) & \# params & mean $\downarrow$ & std $\downarrow$ & match(\%) $\uparrow$\\
    \hline\hline
     1 &0.458& 1.71M & \better{27.47} & 27.83 & \better{48.75}\\
     2 &0.458& 2.09M & 27.75 & 27.32 & 45.42\\
     1 & 0.916 & 1.44M & \best{21.17} & \best{25.49} &  \best{57.50}\\
     2 & 0.916 & 1.72M &  36.40 & \better{25.66} & 20.42\\
     \hline
    \end{tabular}
    }
    \label{tab:ablation}
\end{table}

\section{Conclusion and Future Work}
% \wei{Too many paragraphs. Squeeze them into one.}

We present a learning-based system to generate local BEV images combined with NCC for ground vehicle localization in GNSS-denied off-road environments. Our system incorporates the deformable attention module with BEVFormer for a multi-view camera sensor setting, followed by a novel rendering head to generate high-precision BEV images to enable downstream localization task.

To enhance our ground vehicle localization system for operation across different seasons, future research will focus on improving the network's ability to learn and generalize features from varied seasonal landscapes. This is essential for deploying our system in real-world scenarios where environmental conditions fluctuate significantly over the year. Additionally, we aim to advance the fidelity of BEV image generation by incorporating techniques such as the diffusion module, inspired by the diffusion transformer~\cite{peebles2023scalable}. This enhancement is expected to refine the detail and precision of the BEV images, thus enriching contextual data for more accurate vehicle localization.

\begin{figure}[ht]
    \centering
    \resizebox{\columnwidth}{!}{
    \begin{tabular}{@{}c@{\hspace{1mm}}c@{}}
    \includegraphics[width=0.48\columnwidth]{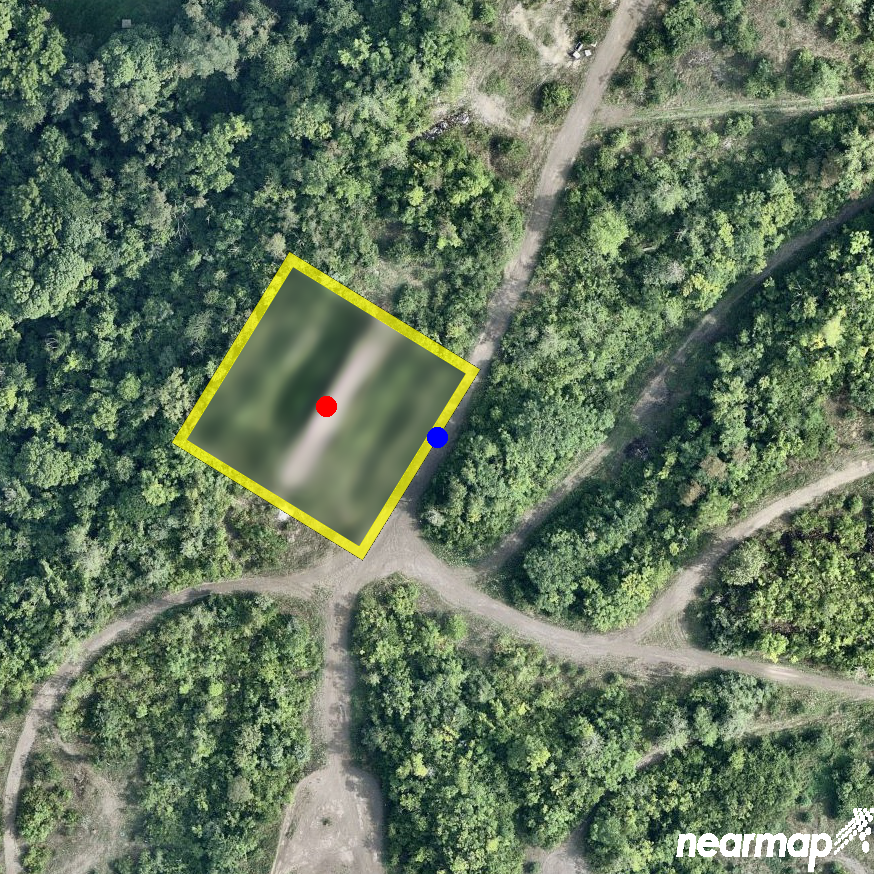} &
    \includegraphics[width=0.48\columnwidth]{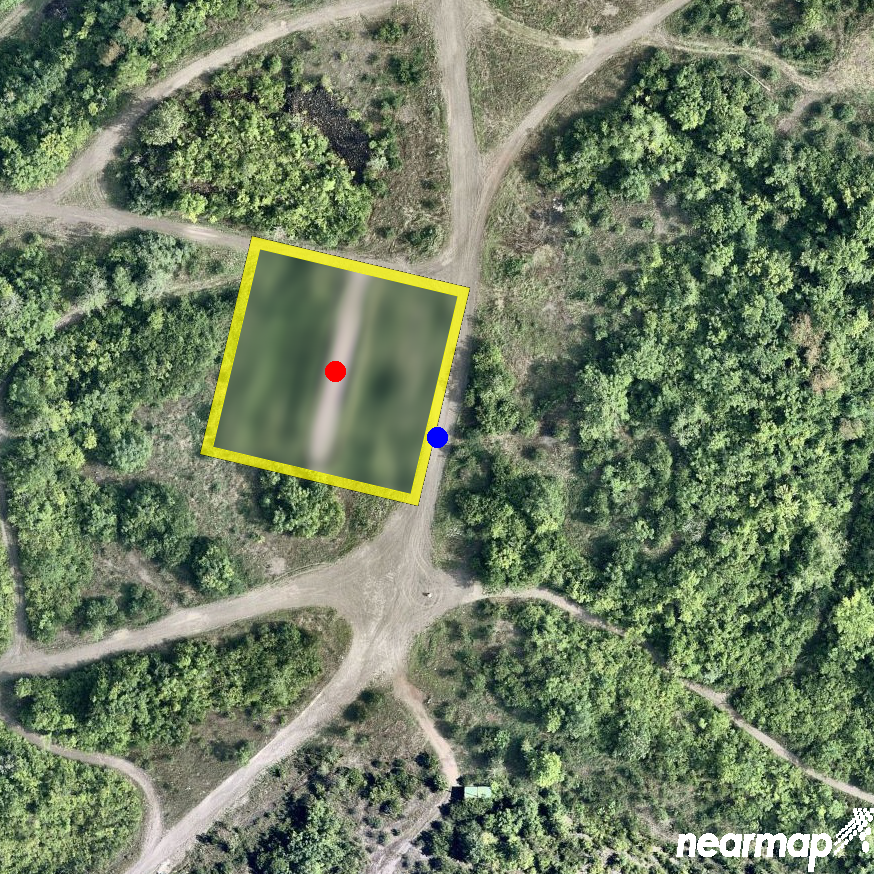}
    \end{tabular}
    }
    \caption{Examples of failure cases due to the uniform weighting of NCC.}
    \label{fig:failure}
\end{figure}

Further improvements will also explore the integration of temporal features to accumulate historical data more effectively, addressing current limitations caused by projection adjustments and vehicle pose changes. Moreover, explorations can be made on removing dependence on GPS information for training by leveraging local state estimates from VIO. Furthermore, a sophisticated approach to incorporate data from previous frames could significantly improve rendering quality and system performance. In addition, a transition from classic template matching to learnable template matching for vehicle positioning is anticipated to overcome the limitation of NCC's uniform pixel weighting, as shown in Fig.~\ref{fig:failure}, and to enable the system to prioritize strategically significant areas, potentially elevating the accuracy of vehicle registration in challenging environments.

\section*{Acknowledgement}
The authors thank Kaicheng Yu for the fruitful discussion and valuable suggestions, and members of the National Robotics Engineering Center (NREC) for helping with data collection.

\bibliographystyle{IEEEtran}
\bibliography{ref.bib}

\end{document}